\RequirePackage{fix-cm}
\documentclass[twocolumn]{svjour3}          
\smartqed  
\usepackage{graphicx}
\usepackage[misc]{ifsym}
\usepackage{cite}
\usepackage{amsmath,amssymb,amsfonts}
\usepackage{algorithmic}
\usepackage{textcomp}
\usepackage{booktabs}
\usepackage{multirow}
\usepackage{CJKutf8}
\usepackage{xcolor}
\usepackage{caption}
\usepackage[numbers,sort&compress]{natbib}
\journalname{}
\begin{document}

\title{Convolutional herbal prescription building method from multi-scale facial features
}


\author{Huiqiang Liao      \and
        Guihua Wen \and
        Yang Hu \and
        ChangJun Wang
}


\institute{Huqiang Liao \at
              School of Computer Science and Engineering in South China University of Technology, Guangzhou, China \\
              \email{huiqiangliao@163.com}           
           \and
           GuiHua Wen(\ \Letter \ ) \at
              School of Computer Science and Engineering in South China University of Technology, Guangzhou, China \\
              \email{crghwen@scut.edu.cn}
           \and
           Yang Hu \at
           School of Computer Science and Engineering in South China University of Technology, Guangzhou, China \\
           \email{superhy199148@hotmail.com}
           \and
           Changjun Wang \at
           Department of Traditional Chinese Medicine in Guangdong General Hospital, China\\
           \email{gzwchj@126.com}
}

\date{ }

\maketitle

\begin{abstract}
In Traditional Chinese Medicine (TCM), facial features are important basis for diagnosis and treatment. A doctor of TCM can prescribe according to a patient’s physical indicators such as face, tongue, voice, symptoms, pulse. Previous works analyze and generate prescription according to symptoms. However, research work to mine the association between facial features and prescriptions has not been found for the time being. In this work, we try to use deep learning methods to mine the relationship between the patient's face and herbal prescriptions (TCM prescriptions), and propose to construct convolutional neural networks that generate TCM prescriptions according to the patient's face image. It is a novel and challenging job. In order to mine features from different granularities of faces, we design a multi-scale convolutional neural network based on three-grained face, which mines the patient's face information from the organs, local regions, and the entire face. Our experiments show that convolutional neural networks can learn relevant information from face to prescribe, and the multi-scale convolutional neural networks based on three-grained face perform better.
\end{abstract}

\section{Introduction}
\label{sec:introduction}
TCM (Traditional Chinese Medicine) was developed through thousands of years of empirical testing and refinement, and played an important role in health maintenance for the Chinese ancient people \cite{Cheung2011Tcm}. It is a theoretical system that is gradually formed and developed through long-term medical practice. TCM has the advantages of convenience, cheap and low side effects, and is suitable for use in hospitals, even in community hospitals with poor conditions. 

Prescription in TCM consists of a variety of herbs, which is the main way to treat diseases for thousands of years. In the long Chinese history, a lot of prescriptions have been invented to treat diseases and more than 100,000 have been recorded \cite{Qiu2007TraditionalMedicine}. An example of a prescription in Dictionary of Traditional Chinese Medicine Prescriptions is given in Figure 1 \cite{Peng1996DictOfTcm, Yao2018TcmTopicModel}. 

There are four important diagnostic methods in TCM: Observing, Listening, Inquiry, Pulse feeling. Observing understands the state of health or disease through objective observation of all visible signs and effluents of the whole body and part of the body. Face diagnosis is a common method of observing, which can understand the pathological state of various organs in the body by observing changes in facial features \cite{Wang2012TcmInspection}. Face appearance signals information about an individual \cite{Jones2018FaceHealth}. The face is rich with capillaries, which is like a mirror that reflects the physiological pathology of humans. From the view of TCM, the characteristics of the various regions of the face represent the health status of various internal organs of the human body. The doctor can judge the physical condition of the patient by observing the facial features of the patient.

Computer aided diagnosis (CAD) based on artificial intelligence (AI) is an extremely important research field in intelligent healthcare \cite{Chen2017DeepLearningForMedical}. According to a survey, deep learning algorithms, especially convolutional neural networks(CNN), have been widely used in various fields of medical image processing in recent years due to their excellent performances in the field of computer vision, such as disease classification, lesion detection, and substructure segmentation \cite{Litjens2017Survey}. From the 306 papers reviewed in this survey, it is evident that deep learning has pervaded every aspect of medical image analysis \cite{Litjens2017Survey}. End-to-end convolutional neural network training has become a good choice for medical image processing tasks.

However, to the best of our knowledge, there has not been research work that mines the relationship between the patient's face and TCM prescriptions. In realistic TCM, doctors prescribe through features of face, tongue, pulse, voice, and symptoms. Using face images to generate TCM prescription is of great significance to assist doctors in the diagnosis and treatment. Especially for some young doctors, the generated prescriptions can give them some references. It can recommend prescriptions to doctors. After making some modifications, doctors can apply them to practice. It saves treatment costs compared to directly prescribe from scratch and improve the efficiency of the doctor's prescribing. A large number of data samples can be used to learn the relevant information of patient's face and TCM prescription. Learning to how to prescribe through the patient’s diagnosis data can provide a reference for TCM doctors to observe and diagnose patients.

In this paper, we propose to use deep learning (convolutional neural network) to prescribe (TCM prescriptions) based on the patient’s face image. The main work is as follows:

1. A conventional convolutional neural network was designed to encode the patient’ face image features and generate TCM prescriptions.

2. Considering different facial organs(eyes, nose, mouth) and regions(cheeks, chin) represent the status of internal organs(heart, liver, spleen, lungs and kidney) in various parts of the human body, a multi-scale convolutional neural network based on three-grained face is proposed to extract feature of facial organs, facial regions and entire face to learn to generate TCM prescriptions.

3. We conduct experiments to verify the effectiveness of convolutional neural network for face feature encoding and prescription generation.

The rest of this paper is organized as follows. In Section 2, we discuss some related work on TCM prescriptions and medical image processing. Section 3 illustrates task description and methodology. Section 4 elaborates and analysis the experiment results. We have some discussions in Section 5 and conclude this paper in Section 6.

\begin{figure}
	\centerline{\includegraphics[width=\columnwidth]{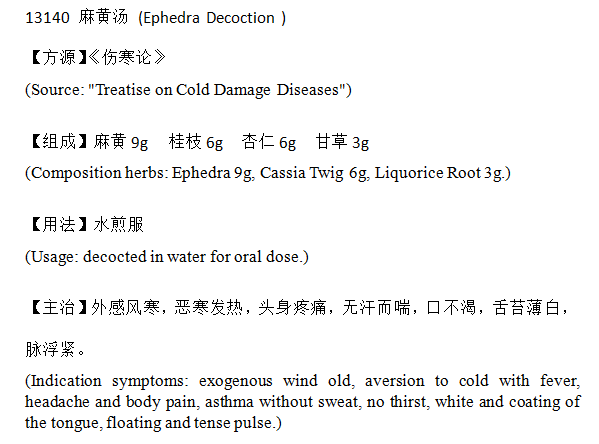}}
	\caption{An example of a prescription in Dictionary of Traditional Chinese Medicine Prescriptions. The Composition herbs are the most important in a prescription, and generally the prescriptions means composition herbs.}
	\label{fig1}
\end{figure}

\section{Related Work}
\textbf{Deep learning in medical image processing.} Deep learning and convolutional neural network have become popular topics in medical image processing. There are already a lot of research works that apply deep learning to medical image processing. In terms of disease classification, there are studies on breast cancer image classification \cite{Bayramoglu2016BreastCancerClassify, Chougrad2018BreastCancerScreening}, lung pattern classification \cite{Anthimopoulos2016LungClassify}, Alzheimer’s disease classification \cite{Hon2017AlzheimerClassify}. In the detection of lesion targets and diseases, there are cancerous tissue recognition \cite{Stanitsas2017CancerRecognize}, detecting cardiovascular disease \cite{Wang2017DetectCardiovascular}, and melanoma recognition \cite{Yu2017MelanomaRecognize}. In the segmentation of organs and substructures, there are studies on skin lesion segmentation \cite{Yuan2017SkinSegment}, microvasculature segmentation of arterioles \cite{Kassim2017MicrovasculatureSegment}, tumor segmentation \cite{Zhao2018BrainTumourSegmantation}. In addition, there are many other applications, such as studies of visual attention of patients with Dementia \cite{Chaabouni2017}, diagnosis of cirrhosis stage \cite{Xu2017CirrhosisDiagnosis}, constructing brain maps \cite{Zhao2017BrainNetworkAtlases}.

\noindent \textbf{TCM prescriptions.} On the other hand, some work has been devoted to the study of TCM prescriptions. Some studies analyzed and explored TCM prescriptions and discovered the regularity \cite{Liu2012TcmDataProcess, Zheng2014Prescription, Xie2017Prescription, Zhang2012Prescription}. Some studies used topic model to discover prescribing patterns \cite{Yao2018TcmTopicModel, Yao2015TcmTopicModel}. There are other studies such as TCM medicine classification and recognition \cite{Dehan2014TcmMedicineClassify, Weng2017TcmRecognize}, knowledge graph \cite{Yu2017TcmKnowledgeGraph, Weng2017TcmKnowledgeGraph} for TCM.

In practice, TCM doctors can judge the health of various internal organs of the body by observing the patient's face. Combining other characteristics, doctors can give TCM prescription based on their knowledge. Our work is to try to simulate and learn this process. Using deep learning techniques, we can learn how to prescribe from a large amount of medical data. At the current stage, the medical data from which we learn are the patient’s face images and corresponding prescriptions. The study is of great importance to assist doctors to diagnosis and treat.

\section{Methodology}
\label{sec:methodology}
\subsection{Data Collection and Task Description}
The data set used in our study are collected from cooperative hospitals. After preprocessing, there are 9,653 data pairs totally. Each data pair contains a patient's face image and a corresponding TCM prescription.

All Chinese herbal medicines are included in a unified dictionary $H=\{h_1,h_2,...,h_n\}$. The i-th element $h_i$ in $H$ represents the i-th Chinese herbal medicines, and there are $n$ Chinese herbal medicines. In our dataset, $n$ is 559. Each TCM prescription can be represented by a vector $y=[y_1,y_2,...,y_n]$. The element $y_i$ in $y$ can only be 1 or 0, indicating whether the Chinese herbal medicine is prescribed. Each patient’s face image is represented by a pixel matrix $x$, and the size of $x$ is 224x224x3. $X$ represents all face images in dataset and $Y$ represents all prescriptions in dataset.

The task of this paper is to input a patient's face image (pixel matrix $x$) and output the patient's corresponding prescription $y$. The prescription $y$ is a multi-label vector. In fact, the task is a multi-label learning. Multi-label learning studies the problem where each example is represented by a single instance while associated with a set of labels simultaneously \cite{Zhang2014MultiLabel}. 

\subsection{Construction of conventional Convolutional Neural Network}
Deep convolutional neural networks are widely used in the field of image processing. It can extract potential features from the original pixel matrix with RGB color channels for use in various image tasks such as classification, detection, segmentation. Classical convolutional neural network structures include AlexNet \cite{Krizhevsky2012AlexNet}, VGGNet \cite{Simonyan2015VGG}, GoogleNet \cite{Szegedy2015GoogleNet}, ResNets \cite{He2016ResNet, Xie2017ResNeXt, Zagoruyko2016wrn}, DenseNet \cite{Huang2017DenseNet} and SENet \cite{Hu2018seNet}.

The convolutional neural network used for prescription generation is composed of several convolutional modules and fully connected layers. Each convolutional module includes a convolutional layer and a pooling layer. In order to extract features from the image, the convolutional layer uses some convolutional kernels to scan image matrix to reconstruct a feature map $C$. A convolutional kernel is a weight matrix. We use $K$ to represent it. The above operation can be abstracted as a function with the relu \cite{Glorot2011Relu} activation function:

\begin{equation}C(x,K)=ReLU(Conv(x,K)).\label{eq}\end{equation}

In order to extract more important features and reduce the computational complexity, the max-pooling layer is used to downsample the feature map $C$, which can be represented by the following function (the parameters of max-pooling layer are omitted):

\begin{equation}\hat{C}(x,K)=Max(C(x,K)).\label{eq}\end{equation}

Three consecutive convolution and pooling operations can be abstracted into the following function:

\begin{equation}\hat{C}^{3}(x,K)=\hat{C}(\hat{C}(\hat{C}(x,K))).\label{eq}\end{equation}

In order to encode features, several fully connected layers are usually connected to the end of several convolution modules. The weight parameters of the fully connected layer layer are denoted by $W1$. An operation of fully connected layer (with a relu activated function) can be abstracted as the following function:

\begin{equation}f(\hat{C}^{3},W1)=ReLU(FC(\hat{C}^{3},W1)).\label{eq}\end{equation}

The last layer is the output layer, which is a fully connected layer with sigmoid activation function. The weight is represented by $W2$. It outputs the probability of whether each Chinese herbal medicine is prescribed, which can be abstracted as the following function:

\begin{equation}
\begin{split}
P(x,\Theta )&=sigmoid(FC(f,W2))\\
&=[P(h_1|x,\Theta ),...,P(h_n|x,\Theta)].
\end{split}
\label{eq}
\end{equation}

$\Theta =\{K,W1,W2\}$ is the set of all parameters, and the convolutional kernel $K$ for each convolutional operation described above is different.

The loss function of the convolutional neural network is designed as the average value of multiple cross-entropy. Each cross-entropy measures the difference between the probability of prescribing of each Chinese herbal medicine $P(h_i|x,\Theta )$ and actual output $y_i$. The neural network minimizes the loss function by optimizing all parameters using stochastic gradient descent \cite{Bottou2012SGD}, which can be abstracted as the following functions($m$ is the size of the dataset):

\begin{equation}
\begin{split}
J(\Theta, x)=&\frac{1}{n}\sum \limits_{i=1}^{n}[-y_ilog(P(h_i|x,\Theta))-\\
&(1-y_i)log(1-P(h_i|x,\Theta))];\\
\end{split}
\label{eq}
\end{equation}

\begin{equation}
\Theta^* = arg \mathop {min }\limits_{\Theta}\frac{1}{m}\sum \limits_{j=1}^{m}J(\Theta, x_j).
\label{eq}
\end{equation}

The structure of the convolutional neural network is shown in Figure 2. It contains three convolution modules for extracting features, a fully connected layer for coding features, and the final output layer. All the sizes of convolution kernels are 3x3. The input of the network is the face image matrix $x$ of the patient, and the size is 224x224x3. The number of elements in the output layer is $n$, the size of the Chinese herbal medicine dictionary $H$, and each unit represents the probability that a certain Chinese herbal medicine is prescribed. The number of dimensions of the real output $y$ is $n$, and each value is 0 or 1, indicating whether to prescribe. The loss is the average cross-entropy loss calculated from the network output $P$ and the real output $Y$. $P$ contains the probabilities of being prescribed for all Chinese herbal medicine in dictionary $H$. Finally, according to dictionary $H$, a final prescription is obtained by sampling from $P$ through a probability threshold t. 


\begin{figure}
	\centerline{\includegraphics[width=0.6\columnwidth]{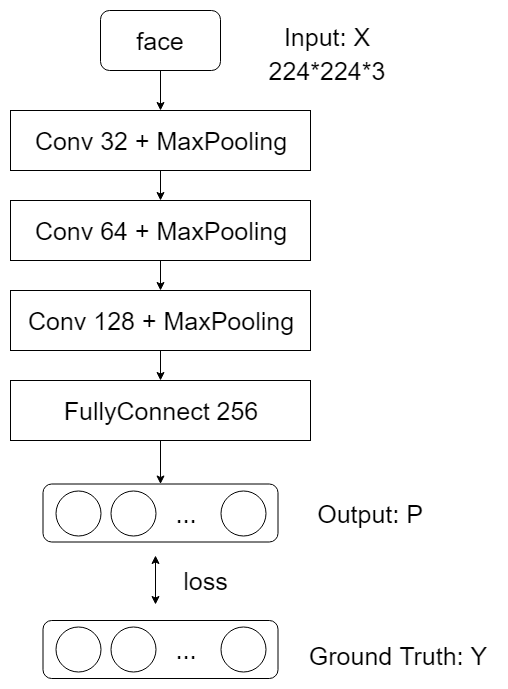}}
	\caption{A conventional convolutional neural network for generating TCM prescriptions}
	\label{fig2}
\end{figure}

\subsection{Construction of multi-scale Convolutional Neural Network based on three-grained face}
Different regions of a face image have different local statistics\cite{Taigman2014DeepFace}. Taigman et al. \cite{Taigman2014DeepFace} use locally connected layers, which like a convolutional layer but every location in the feature map learns a different set of filters, to deal with this problem. However, the use of local layers greatly increases the parameters of the model. Only a large amount of data can support this approach, so instead of doing this, we extract features of different facial regions using different small convolutional networks. 

According to TCM, the characteristics of various regions of the face represent the health of various internal organs of the human body. In order to encode the features of each region of the face more efficiently, the paper proposes a multi-scale convolutional neural network based on three-grained face. The ``three-grained'' refers to the organ block, the local region block, and the face block. Each block extracts characteristics of the face area from different granularities. The organ block includes the left eye, right eye, nose, and mouth. The local region block includes the left cheek, right cheek, and chin. The face block means the entire face. The network is expected to extract and encode more effective facial features from different granularities, thereby improving the effectiveness of prescription generation.   

In the data preprocessing stage, the patient's face is segmented to obtain different region images of the face. An example of different region images after cutting the face \cite{Vidit2010fddb} is given in Figure 3. The sizes of different regions images are reduced. The organ block images $X_{organ}$ includes a left-eye image $X_{o-1}$, a right-eye image $X_{o-2}$, a nose image $X_{o-3}$, a mouth image $X_{o-4}$, and their sizes are 56x56x3. The local region block images $X_{region}$ includes a left cheek image $X_{r-1}$, a right cheek image $X_{r-2}$ and a chin image $X_{r-3}$, and their sizes are 112x112x3. The face block means to the entire face $X_{face}$, and the size of face image is 224x224x3.

\begin{figure}
	\centerline{\includegraphics[width=1\columnwidth]{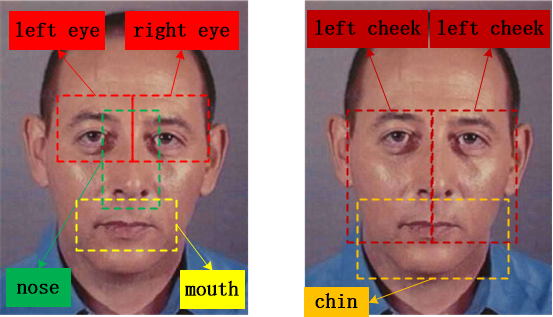}}
	\caption{Different organ and local region images of the segmented face. \textbf{Left:} The organ block images. \textbf{Right:} The local region block images.}
	\label{fig3}
\end{figure}

\subsubsection{Extracting feature of facial organ}
Firstly, feature extraction is performed on the organ block. After convolution of four organ block images, concatenate the four feature maps. The operation can be abstracted as the following functions:

\begin{equation}
\begin{split}
C_{o-i}=C(X_{o-i},K), i=\{1,2,3,4\};\label{eq}
\end{split}
\end{equation}

\begin{equation}
\begin{split}
Concat_{o}=Concat(&C_{o-1},C_{o-2},C_{o-3},C_{o-4}).
\end{split}
\label{eq}
\end{equation}

In the field of computer vision applications, there is often not enough data, and the overfitting of models easily occur. Usually, dropout \cite{Srivastava2014Dropout} is used to prevent overfitting. Dropout randomly discards neural units during training phase. This prevents units from co-adapting too much and force the network to learn more robust features. It reduces the size of the network during the training phase and gets a number of more streamlined networks that have similar integration effects \cite{Srivastava2014Dropout}. After dropout the above feature map $Concat_{o}$, a convolution operation is performed again to obtain a feature map $C_{o}$, which extracts features of organ block. The above operations can be abstracted as the following function:

\begin{equation}
C_{o}=C(Concat_{o},K).
\label{eq}
\end{equation}

\subsubsection{Extracting feature of facial local region}
Secondly, feature extraction is performed on the local region block. After convolution and max-pooling of the three local region block images, concatenate the three local region block feature maps together with the feature map extracted by the organ block. The above operation can be abstracted as the following functions:

\begin{equation}
C_{r-i}=\hat{C}(X_{r-i},K),i=\{1,2,3\};
\label{eq}
\end{equation}

\begin{equation}
\begin{split}
Concat_{o\_r}=Concat(&C_{o},C_{r-1},C_{r-2},C_{r-3}).
\end{split}
\label{eq}
\end{equation}

After dropout the above feature map $Concat_{o\_r}$, convolution and max-pooling operations are performed to extract features to obtain a feature map $C_{o\_r}$ (fuses the features of the organ block and local region block). The above operation can be abstracted as the following function:

\begin{equation}
C_{o\_r}=\hat{C}(Concat_{o\_r},K);
\label{eq}
\end{equation}

\subsubsection{Extracting feature of entire face}
Finally, feature extraction is performed on the face block. After several convolution and max-pooling of the entire face, concatenate the face block feature map together with the feature map $C_{o\_r}$. The above operation can be abstracted as the following function:

\begin{equation}C_{face}=\hat{C}(\hat{C}(\hat{C}(X_{face},K)));\label{eq}\end{equation}
\begin{equation}Concat_{o\_r\_f}=Concat(C_{o\_r},C_{face}).\label{eq}\end{equation}

After dropout the above feature map $Concat_{o\_r\_f}$, two fully connected layers are used to encode feature to get the final features (fuse the features of organ block, region block and face block). The above operation can be abstracted as the following function, where $W3$ and $W4$ are the weights of the fully connected layers.

\begin{equation}C_{o\_r\_f}=f(f(Concat_{o\_r\_f},W3),W4)\label{eq}\end{equation}

\subsubsection{Training based on three-grained face features (organ, local region, entire face)}

The convolutional neural network has three output layers. The first output $P_{organ}$ uses the feature map $C_{o}$, which extracts the features of organ block, to predict. The second output $P_{region}$ uses the feature map $C_{o\_r}$, which extracts the features of organ block and region block, to predict. The third output $P_{face}$ uses the final feature $C_{o\_r\_f}$, which extracts the features of organ block, region block and face block, to predict. The above operation can be abstracted as the following function, where $W_{o1}$, $W_{o2}$ and $W_{o3}$ represent the weights of output layers.

\begin{equation}
\begin{split}
P_{organ}=f(C_{o},W_{o1})
\end{split}
\label{eq}
\end{equation}

\begin{equation}
\begin{split}
P_{region}=f(C_{o\_r},W_{o2})
\end{split}
\label{eq}
\end{equation}

\begin{equation}
\begin{split}
P_{face}=f(C_{o\_r\_f},W_{o3})
\end{split}
\label{eq}
\end{equation}

$P_{organ}$, $P_{region}$, and $P_{face}$ denote the probabilities of being prescribed for all Chinese herbal medicine in dictionary $H$. Among them, $P_{face}$ is the main output of the neural network, which is the decision output of the final generation. $P_{organ}$ and $P_{region}$ are auxiliary outputs, which are used to assist the training of the entire network. The final loss is addition of three losses, which are calculated by $P_{organ}$, $P_{region}$, and $P_{face}$ and the real output $Y$. We use stochastic gradient descent to optimize the parameters so that the final loss is minimized. The loss functions are as follow, where $\Theta$ denote the set of all parameters of the neural network and $n$ means the dimension of each real prescription $y$.

\begin{equation}
\begin{split}
J1(\Theta )=\frac{1}{n}\sum \limits_{i=1}^{n}[&-Y_ilog(P_{organ})-\\&(1-Y_i)log(1-P_{organ})]
\end{split}
\label{eq}
\end{equation}

\begin{equation}
\begin{split}
J2(\Theta )=\frac{1}{n}\sum \limits_{i=1}^{n}[&-Y_ilog(P_{region})-\\&(1-Y_i)log(1-P_{region})]
\end{split}
\label{eq}
\end{equation}

\begin{equation}
\begin{split}
J3(\Theta )=\frac{1}{n}\sum \limits_{i=1}^{n}[&-Y_ilog(P_{face})-\\&(1-Y_i)log(1-P_{face})]
\end{split}
\label{eq}
\end{equation}

\begin{equation}
\Theta^* = arg \mathop {min }\limits_{\Theta}{J1(\Theta)+J2(\Theta)+J3(\Theta)}.
\label{eq}
\end{equation}

The multi-scale convolutional neural network based on three-grained face structure is shown in Figure 4, in which the sizes of the input organ block images are 56x56x3, and the sizes of the input region block images are 112x112x3, the size of the input face block image is 224x224x3. All the sizes of convolution kernels are 3x3. 

The network is divided into three parts. The first part extracts the features of organ block to obtain output $P_{organ}$. The second part extracts the features of region block and then merges them with the features of the organ block to continue to extract feature to get the output $P_{region}$. The third part extracts the features of face block and then merges them with the features of the organ block and region block to continue to extract feature to get the output $P_{face}$. The three outputs denote the probabilities of being prescribed for all Chinese herbal medicine in dictionary $H$. The loss used to train the entire network is addition of three losses, which are calculated by $P_{organ}$, $P_{region}$, $P_{face}$ and the real output $Y$. Finally, the final generated prescription is obtained by sampling from the output $P_{face}$ through the probability threshold $t$. 

\begin{figure*}
	\centerline{\includegraphics[width=1.4\columnwidth]{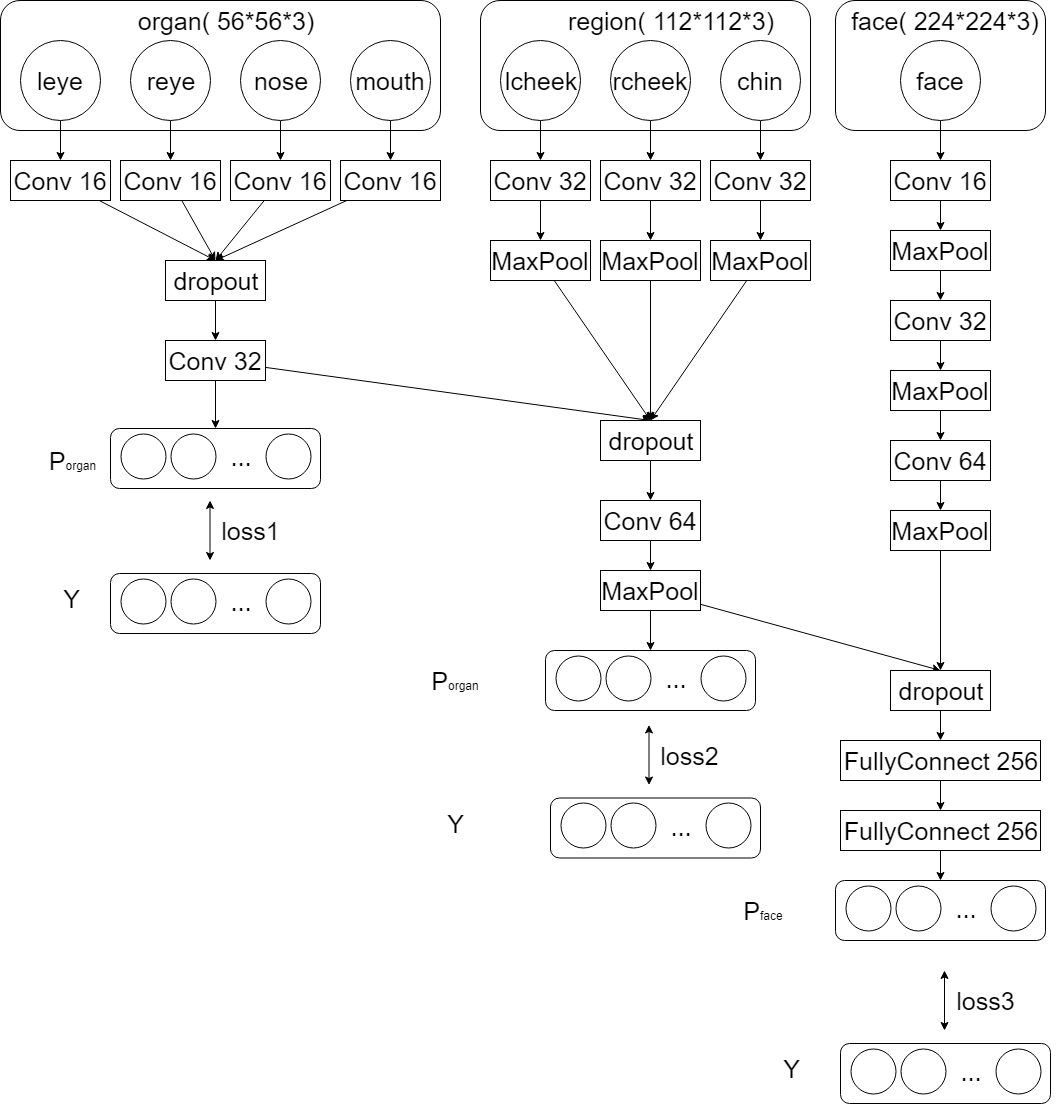}}
	\caption{A multi-scale convolutional neural network based on three-grained face for generating TCM prescriptions. }
	\label{fig4}
\end{figure*}

\subsection{Data augmentation}
In the real world, patient's medical data is precious and difficult to collect. Therefore, the data collected from the patient's faces and prescriptions are very limited. Due to the limited data set, it is easy to cause the model to overfit, which is one reason for not choosing an overly complex network. Data augmentation is an effective way to cope with not enough data. It can reduce overfitting of the model and improve the model's predictive performance.

In order to make full use of limited data, data augmentation is performed. The ``data augmentation'' randomly extracts some of the original patient’s face images, then randomly transforms the images (such as rotation, zoom) and then saves the image as a new patient's face image. The original patient’s prescription are used as the prescription labels of the new patient’s face image. Data augmentation can increase the size and diversity of the data set. The sample size of the original data set is 9653. After data augmentation, the data set size increases to 18,463. Some parameters used in data augmentation are shown in Table 1.

\begin{table}
	\caption{Parameters of data augmentation}
	\label{table}
	\setlength{\tabcolsep}{5pt}
	\linespread{1.5}\selectfont
	\begin{tabular}{p{75pt}p{25pt}p{115pt}}
		\toprule[2pt]
		Parameters           & Values  & Explanations \\
		\midrule[1pt]
		rotation\_range      & 25      & A value in (0-180) within which images are randomly rotated \\
		width\_shift\_range  & 0.05    & Translate horizontally by a certain percentage of width \\
		height\_shift\_range & 0.05    & Translate vertically by a certain percentage of height \\
		zoom\_range          & 0.2     & The zoom range is [1-0.2, 1+0.2] \\
		horizontal\_flip     & True    & A flag for randomly flipping half of the images \\
		\bottomrule[2pt]
	\end{tabular}
	\label{tab1}
\end{table}

\section{Experiment}
\label{sec:experiment}

\subsection{Dataset}
``Face image - TCM prescription'' dataset is collected from some cooperative hospitals. Due to the limited collection conditions, the collected raw data have a certain noise. For example, there are different medicine names but exactly they are the same medicine. After some preprocessing, the experimental dataset is obtained. The size of experimental dataset $D_{origin}$ is 9653. After data augmentation, the size of dataset is increased to 18463, and the dataset is denoted as $D_{aug}$. In order to train multi-scale convolutional neural network based on three-grained face, the face images are segmented into different face areas: eyes, nose, mouth, cheeks, and chin. The specific description of the dataset is shown in Table 2 and Table 3.

In order to enhance the accuracy and persuasiveness of the experimental results, we use 5-fold cross-validation method to train and evaluate model: repeatedly performs training for five times and 500 samples are taken as test set for each time(conventional approach should divide data into five equal parts, each equal part is taken as the test set for each time. Only 500 samples are taken as test set for each time here due to the limited dataset). The 500 test samples taken for each time do not overlap. The average of five evaluation results is used as the final evaluation result.

\begin{table}
	\caption{face images information}
	\label{table}
	\setlength{\tabcolsep}{5pt}
	\linespread{1.5}\selectfont
	\begin{tabular}{p{175pt}p{40pt}}
		\toprule[2pt]
		Parameters                                                       & values \\
		\midrule[1pt]
		number of face images                                            & 9653 \\
		number of face images after data augmentation                    & 18463 \\
		size of face images                                              & 224x224x3 \\
		size of local region block images(left cheek, right cheek, chin) & 112x112x3 \\
		size of organ block images(left-eye, right-eye, nose, mouth)     & 56x56x3 \\
		size of test set selected in 5-fold cross-validation             & 500 \\
		\bottomrule[2pt]
	\end{tabular}
	\label{tab2}
\end{table}

\begin{table}
	\caption{prescription information}
	\label{table}
	\setlength{\tabcolsep}{5pt}
	\linespread{1.5}\selectfont
	\begin{tabular}{p{190pt}p{20pt}}
		\toprule[2pt]
		Parameters                                                                      & values \\
		\midrule[1pt]
		number of prescriptions                                                         & 9653 \\
		number of Chinese herbal medicine species                                       & 559 \\
		maximum length of TCM prescriptions           & 28 \\
		minimum length of TCM prescriptions           & 2 \\
		average length of TCM prescriptions           & 14 \\
		Average times each Chinese herbal medicine appears in prescriptions                                 & 240 \\
		Proportion of Chinese herbal medicine that appear in more than 100 prescrptions & 36\% \\
		\bottomrule[2pt]
	\end{tabular}
	\label{tab3}
\end{table}

\subsection{Experimental setup}
According to conventional convolutional neural network, multi-scale  convolutional neural network based on three-grained face, and data augmentation, five models are run for TCM prescription generation, briefly described as follows: 

Random forest (baseline): Random forest \cite{Breiman2001RandomForest} classifier is used to generate TCM prescriptions. The features are face images matrix and the labels are multi-label vectors representing the TCM prescriptions. 

Conventional $CNN$: Construct a CNN as described in section 3.2 to train according to face images and TCM prescriptions to obtain a model for generating TCM prescriptions. The experimental data set used is $D_{orgin}$.

Conventional $CNN_{aug}$: The method is the same as conventional $CNN$, but the experimental data set used is $D_{aug}$.

Multi-scale $CNN$ based on three-grained face: Construct a CNN as described in section 3.3 to train to obtain a model for generating TCM prescriptions according to images of different face regions and TCM prescriptions. The experimental data set used is $D_{orgin}$. 

Multi-scale $CNN_{aug}$ based on three-grained face: The method is the same as multi-scale $CNN$ based on three-grained face, but the experimental data set used is $D_{aug}$.

The structure and some parameters of the conventional $CNN$ and multi-scale $CNN$ based on three-grained face have been described in section 3.2 and section 3.3. The more specific parameters are shown in Table 4. The optimization algorithm is SGD (stochastic gradient descent), and learning rate decay is 1e-6, and momentum is 0.9.

\begin{table}
	\caption{parameters of neural networks}
	\label{table}
	\setlength{\tabcolsep}{5pt}
	\linespread{1.5}\selectfont
	\begin{tabular}{p{175pt}p{40pt}}
		\toprule[2pt]
		Parameters                                                         & values \\
		\midrule[1pt]
		the size of convolutional kernel                                   & 3x3 \\
		numbers of convolutional kernel in conventional CNN                & 32,64,128 \\
		numbers of convolutional kernel in multi-scale CNN based on three-grained face & 16,32,64 \\
		the size of max-pooling                                            & 2x2 \\
		dropout rate                                                       & 0.4 \\
		number of neurons in the fully connected layer                     & 256 \\
		activation of convolutional layer and fully connected layer        & relu \\
		number of output layer                                             & 559 \\
		activation of output layer                                         & sigmoid \\
		learning rate                                                      & 0.01 \\
		batch size                                                         & 64 \\
		number of epochs                                                   & 300 \\
		\bottomrule[2pt]
	\end{tabular}
	\label{tab4}
\end{table}

\subsection{Evaluation metrics}
In order to measure the similarity between the generated TCM prescription and the actual TCM prescription, the indicators precision, recall, and f-score are set as shown in the following formulas. $n\_true_i$ denotes the number of Chinese herbal medicine appearing in both the i-th generated prescription and the i-th real prescription. $n\_predict_i$ denotes the number of Chinese herbal medicine appearing in the i-th generated prescription. $n\_real_i$ denotes the number of Chinese herbal medicine appearing in the i-th real prescription. $precision_i$ measures the how the Chinese herbal medicines are precise in generated prescription, and $recall_i$ measures the how the Chinese herbal medicines are complete in generated prescription. $f1\_score_i$ ($f\_score_i$) is the harmonic mean of $precision_i$ and $recall_i$, neutralizing these two indicators.

\begin{equation}
prescription_i = \frac{n\_true_i}{n\_predict_i}
\label{eq}
\end{equation}

\begin{equation}
recall_i = \frac{n\_true_i}{n\_real_i}
\label{eq}
\end{equation}

\begin{equation}
f1\_score_i = \frac{2*prescription_{i}*recall_{i}}{precision_{i}+recall_{i}}
\label{eq}
\end{equation}

The indicators are calculated for each sample generated by the model, and then averaged to obtain the indicators used to evaluate the quality of the model:

\begin{equation}
precision=\frac{1}{m}\sum \limits_{i=1}^{m}{prescision_i},
\label{eq}
\end{equation}

\begin{equation}
recall=\frac{1}{m}\sum \limits_{i=1}^{m}{recall_i},
\label{eq}
\end{equation}

\begin{equation}
f1\_score=\frac{1}{m}\sum \limits_{i=1}^{m}{f1\_score_i},
\label{eq}
\end{equation}

where m is the size of the dataset. Test set is used to evaluate the model and the size is 500. 

For each example $x_i$, $f1\_score_i$ is the  harmonic mean of $precision_i$ and $recall_i$. But note that $precision$, $recall$, $f1\_score$ are averages, so $f1\_score$ is not the harmonic mean of $precision$ and $recall$.

\subsection{Results and Analysis}

\subsubsection{Training process}
In order to prevent overfitting, the model uses data augmentation, dropout methods. In addition, the strategy ``EarlyStopping'' is also used in the experiment. During training, a certain percentage of data is divided from the training set as a validation set for training observations. The proportion used in the experiment is 0.1. The 10\% of training data is used as a validation set that does not participate in training. During the training process, observe the loss of the model on the validation set. After the validation set loss is no longer declining, wait for a certain number of iterations (we use 10 in the experiment) to stop the training. This can prevent the model from overfitting the training set and make a better prediction of the test set.

Take one of the training results in the 5-fold cross-validation. The changes of the training set and the validation set's loss during the training process are shown in Figure 5 and Figure 6. It can be seen that although the number of epochs is 300(ensure sufficient number of iterations), training is usually stopped at about 30-70 iterations, and the later iterations overfit in the training set. With data augmentation, compared to the conventional CNN, the relative gap between the loss of the training set and the validation set in multi-scale CNN based on three-grained face is smaller, which indicates that the generalization ability of the multi-scale CNN based on three-grained face is relatively high.

\begin{figure*}
	\centerline{\includegraphics[width=1.9\columnwidth]{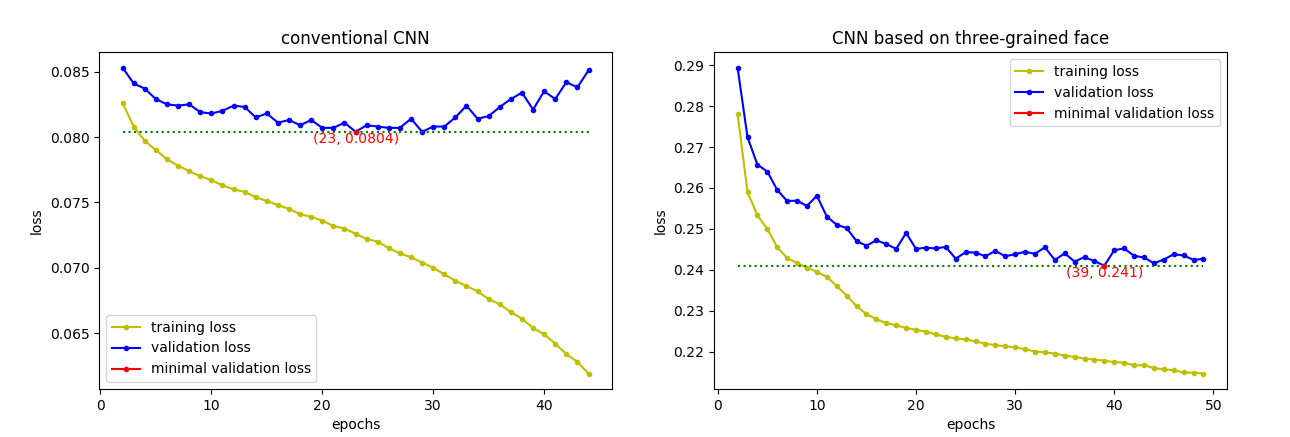}}
	\caption{Learning curve of conventional CNN and multi-scale CNN based on three-grained face prescriptions}
	\label{fig5}
\end{figure*}

\begin{figure*}
	\centerline{\includegraphics[width=2\columnwidth]{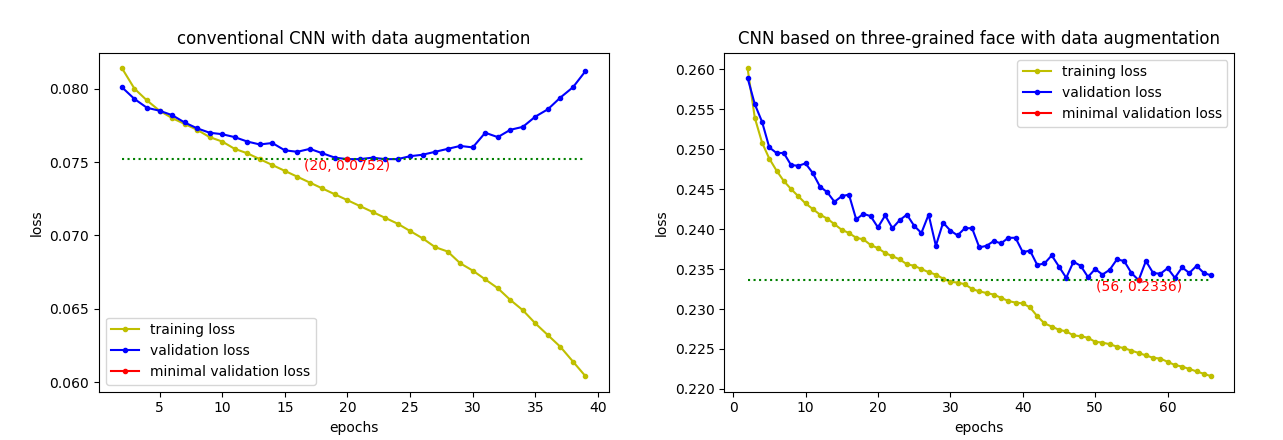}}
	\caption{Learning curve of conventional CNN and multi-scale CNN based on three-grained face (with data augmentation)}
	\label{fig6}
\end{figure*}

\subsubsection{Influence of threshold parameter}
From the final output of the neural network, a series of probability values can be obtained. Finally, the outputs are 559 neurons, representing 559 Chinese herbal medicines. Finally, 559 corresponding probability values are obtained. The final prescription is predicted based on a threshold value t. The Chinese herbal medicine is prescribed if the probability of the Chinese herbal medicine is more than t. 

One general choice for threshold is 0.5. Furthermore, when all the unseen instances in the test set are available, the threshold can be set to minimize the difference on certain multi-label indicator between the training set and test set \cite{Zhang2014MultiLabel}. As shown in Figure 6 and Figure 7, setting different thresholds, the final evaluation results will be different (the results in the figure are the average results of 5-fold cross validation). When a larger threshold is set, a higher precision will be obtained because the prescription generated by the model try to be as precise as possible without errors, and it prefer to give fewer medicines to prevent errors. When a smaller threshold is set, a higher recall is achieved because the prescription generated by the model attempted to be as complete as possible and at the expense of a certain of precision. The ``f1\_score'' is the harmonic mean of precision and recall, which neutralizes the accuracy and completeness. Note that the f1\_score shown in the experimental data is not harmonic mean of precision and recall, because the f1\_score is an average. We choose 0.25 as the final threshold, because at this time the value of f1\_score is high relatively, and the difference between precision and recall is small, which can ensure high precision and recall simultaneously.

\begin{figure*}
	\centerline{\includegraphics[width=1.9\columnwidth]{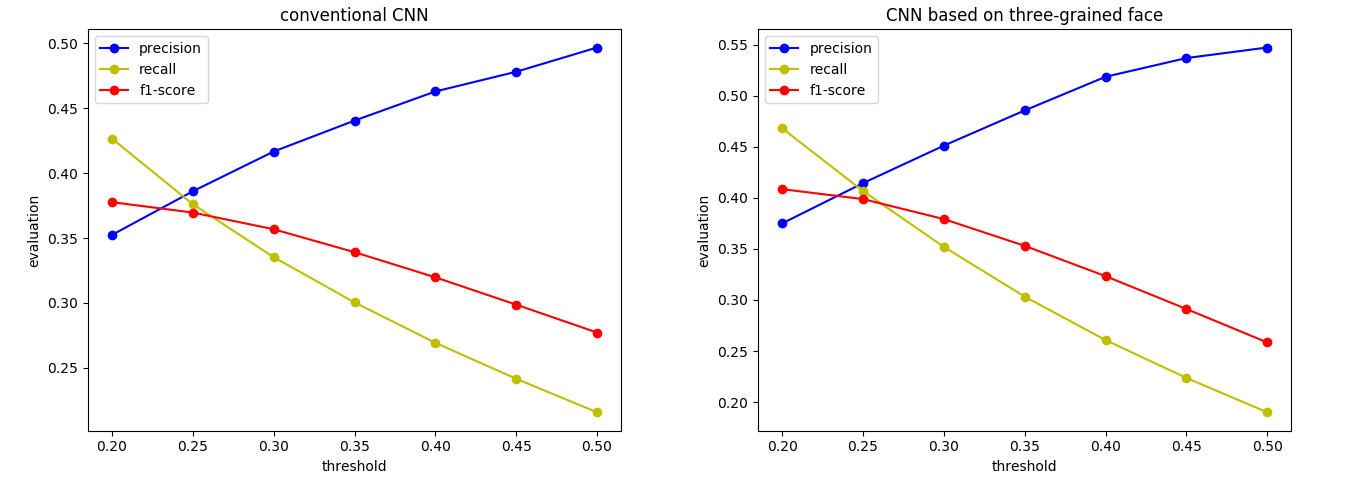}}
	\caption{Influence of threshold on conventional CNN and multi-scale CNN based on three-grained face}
	\label{fig7}
\end{figure*}

\begin{figure*}
	\centerline{\includegraphics[width=1.9\columnwidth]{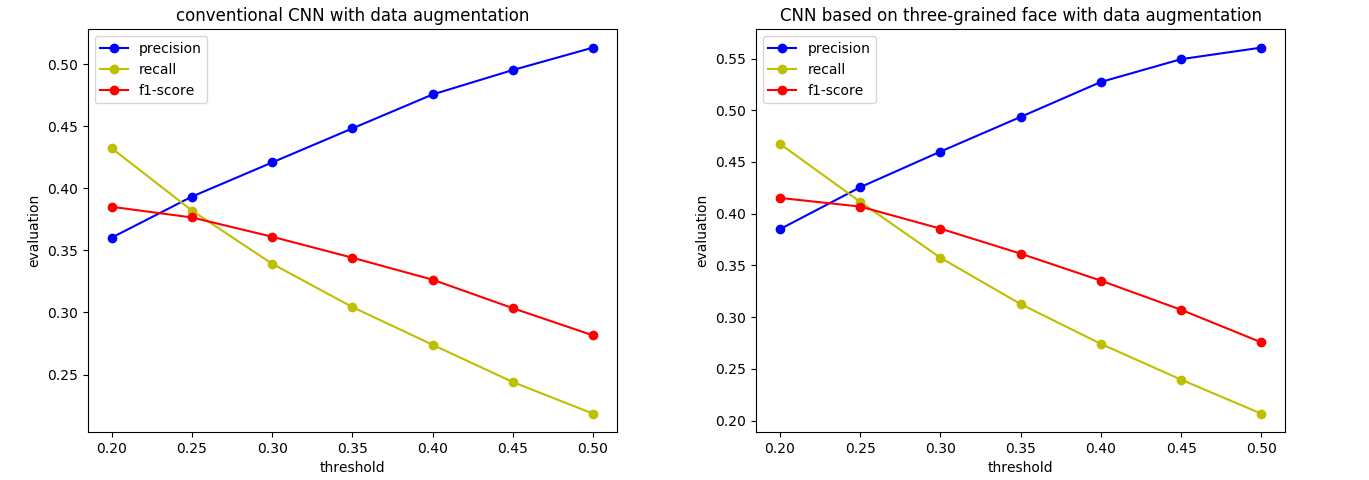}}
	\caption{Influence of threshold on conventional CNN and multi-scale CNN based on three-grained face (with augmentation)}
	\label{fig8}
\end{figure*}

\subsubsection{Performance Comparison}
The experimental results of the five models are shown in Table 5. In order to enhance accuracy and persuasiveness of results, the evaluation results are averaged by 5 results, calculated by 5-fold cross validation methods. The values after ``$\pm$'' indicate the standard deviation of the 5 results. 

Random forest is a ensemble learning technique, which should give good performances. However, it can be seen from the experimental results that the other four models improve the performances compared to the baseline classifier random forest, indicating that the convolutional neural network is better than the random forest in this task. The neural network can extract and represent useful features from large and complex data. There are a large number of original image features that need to be extracted and represented on the task, so using a convolutional neural network for image processing to build a model is a better choice. 

The performances of conventional $CNN_{aug}$ are better than conventional $CNN$, and the performance of multi-scale $CNN_{aug}$ based on three-grained face are better than the multi-scale $CNN$ based on three-grained face. It can be seen that after using data augmentation, the models perform better because using data augmentation increases the size and diversity of the data, allowing the convolutional neural network to learn more knowledge when training. It can reduce overfitting of model. 

The performances of multi-scale $CNN$ based on three-grained face are better than conventional $CNN$, and the performances of multi-scale $CNN_{aug}$ based on three-grained face are better than the conventional $CNN_{aug}$. A reasonable explanation for this result is that the multi-scale $CNN$ based on three-grained face extracts features from different granularities(organs, local regions, and the entire face), and it can extract and utilize local features and global features more effectively.

As shown in Table 7, three samples were taken to show the actual predicted results. For each example, patient's face image and corresponding prescriptions are shown. It can be seen that the results of the model prediction have certain similarities with the actual prescriptions, which shows that the model has indeed learned something. In the four models, the results of multi-scale $CNN_{aug}$ based on three-grained face(we omit the ``multi-scale'' just for neat alignment in table 7) are the most precise and complete. It can be seen that for common Chinese herbal medicines, the prediction of the model will be more accurate, such as Radix Glycyrrhizae and Poria Cocos. For some unusual Chinese herbal medicines, the model cannot accurately predict, such as Perilla Stem and Curcuma Zedoary. A reasonable explanation for this phenomenon is that common Chinese herbal medicines always appear in the training samples, and the model can learn more useful distinguishable features from a large number of training data. However, it is rarely used for the unusual Chinese herbal medicines, which only occasionally used by a few patients. With a small amount of data, the model is difficult to learn. The model cannot find distinguishable features. 

\begin{table*}
	\caption{experimental performances of different models}
	\label{table}
	\setlength{\tabcolsep}{1pt}
	\linespread{1.5}\selectfont
	\begin{tabular}{p{210pt}p{90pt}p{90pt}p{90pt}}
		\toprule[2pt]
		model  & precision(\%) & recall(\%)  & f1-score(\%) \\
		\midrule[1pt]
		random forest - baseline            & 37.25 $\pm$  2.12 & 35.50 $\pm$ 2.97  & 33.07 $\pm$ 1.96\\
		conventional $CNN$                  & 38.60 $\pm$  2.76 & 37.60 $\pm$ 1.53 & 36.69 $\pm$  1.63 \\
		conventional $CNN_{aug}$            & 39.34 $\pm$ 2.56 & 38.21 $\pm$ 1.92 & 37.66 $\pm$ 1.97 \\
		multi-scale $CNN$ based on three-grained face   & 41.44 $\pm$ 2.65 & 40.67 $\pm$ 1.70 & 39.88 $\pm$ 0.73 \\
		multi-scale $CNN_{aug}$ based on three-grained face & \textbf{42.56} $\pm$ 2.48  & \textbf{41.15} $\pm$ 2.01 & \textbf{40.69} $\pm$ 1.67 \\
		\bottomrule[2pt]
	\end{tabular}
	\label{tab5}
\end{table*}

\begin{table*}
	\caption{experimental performances of different image sizes}
	\label{table}
	\setlength{\tabcolsep}{10pt}
	\linespread{1.5}\selectfont
	\begin{tabular}{|p{195pt} | p{17pt}|p{17pt}|p{17pt}|p{25pt}|p{25pt}|  p{25pt}|p{25pt}|}
		\toprule[2pt]
		image size  & 32x32 & 56x56 & 84x84 & 112x112 & 168x168 & 224x224 & average \\
		\midrule[0.5pt]
		precision(\%) of conventional $CNN$                 & 38.97 & 39.42 & 39.27 & 38.83 & 38.18 & 36.69 & \textbf{38.48} \\
		precision(\%) of $CNN$ based on three-grained face  & 37.88 & 38.28 & 38.66 & 39.93 & 40.04 & 39.88 & \textbf{39.36} \\
		\midrule[0.5pt]
		recall(\%) of conventional $CNN$                    & 38.76 & 39.10 & 39.34 & 38.41 & 37.70 & 37.60 & \textbf{38.48} \\
		recall(\%) of $CNN$ based on three-grained face     & 37.97 & 38.37 & 38.70 & 40.16 & 40.59 & 40.67 & \textbf{39.41} \\
		\midrule[0.5pt]
		f1-scores(\%) of conventional $CNN$                 & 38.97 & 39.42 & 39.27 & 38.83 & 38.18 & 36.69 & \textbf{38.61} \\
		f1-scores(\%) of $CNN$ based on three-grained face  & 37.88 & 38.28 & 38.66 & 39.93 & 40.04 & 39.88 & \textbf{39.11} \\
		\bottomrule[2pt]
	\end{tabular}
	\label{tab6}
\end{table*}

\begin{table*}
	\scriptsize
	\caption{real predicted results of different models}
	\label{table}
	\setlength{\tabcolsep}{6pt}
	\linespread{1.3}\selectfont
	\begin{tabular}{p{50pt}|p{120pt}p{310pt}}
		\toprule[2pt]
		\multirow{10}{*}[-80pt]{\raisebox{\height}{\includegraphics[width=50pt]{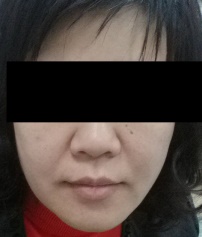}}} & \multirow{2}*{real prescription} & 
		\begin{CJK}{UTF8}{gbsn}
			甘草 白芍 川芎 当归 茯苓 党参 白术 黄连 熟地黄 肉桂 厚朴 白芷 蔓荆子
		\end{CJK}
		\\
		~                  & ~                                                &
		(Radix Glycyrrhizae, Radix Paeoniae Alba, Rhizoma Chuanxiong, Angelica, Poria Cocos, Radix Codonopsis, Macrocephalae Rhizoma, Goldthread, Prepared Rehmannia Root, Cinnamon, Mangnolia Officinalis,  Radix Angelicae Dahuricae, Fructus Viticis)
		\\
		
		\cline{2-3}
		~                  & \multirow{2}*{conventional $CNN$}                 &
		\begin{CJK}{UTF8}{gbsn}
			\textcolor{red}{\textbf{甘草 茯苓 白术}} 浙贝母 砂仁 蜈蚣子
		\end{CJK}
		\\
		~                 & ~                                                 &
		(\textcolor{red}{\textbf{Radix Glycyrrhizae, Poria Cocos, Macrocephalae Rhizoma,}} Thunberg Fritillary Bulb, Fructus Amomi, Scolopendra)
		\\
		
		\cline{2-3}
		~                  & \multirow{2}*{conventional $CNN_{aug}$}           &
		\begin{CJK}{UTF8}{gbsn}
			\textcolor{red}{\textbf{甘草}} 法半夏 \textcolor{red}{\textbf{茯苓 党参 白术}} 山药
		\end{CJK}
		\\
		~                 & ~                                                 &
		(\textcolor{red}{\textbf{Radix Glycyrrhizae,}} Rhizoma Pinellinae Praeparata,  \textcolor{red}{\textbf{Poria Cocos, Radix Codonopsis, Macrocephalae Rhizoma,}} Dioscoreae Rhizoma)
		\\
		
		\cline{2-3}
		~                  & \multirow{2}*{$CNN$ based on three-grained face}  &
		\begin{CJK}{UTF8}{gbsn}
			\textcolor{red}{\textbf{甘草 茯苓 党参 白术}} 蜈蚣
		\end{CJK}
		\\
		~                  & ~                                                 &
		(\textcolor{red}{\textbf{Radix Glycyrrhizae, Poria Cocos, Radix Codonopsis, Macrocephalae Rhizoma,}} Scolopendra)
		\\
		
		\cline{2-3}
		~                  & \multirow{2}*{$CNN_{aug}$ based on three-grained face}  &
		\begin{CJK}{UTF8}{gbsn}
			\textcolor{red}{\textbf{甘草 白芍 茯苓 党参 白术}} 蜈蚣
		\end{CJK}
		\\
		~                  & ~                                                 &
		(\textcolor{red}{\textbf{Radix Glycyrrhizae, Radix Paeoniae Alba, Poria Cocos, Radix Codonopsis, Macrocephalae Rhizoma,}} Scolopendra)
		\\
		
		\midrule[2pt]
		\multirow{10}{*}[-80pt]{\raisebox{\height}{\includegraphics[width=50pt]{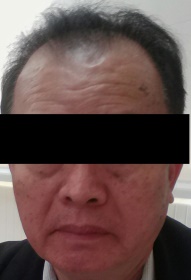}}} & \multirow{2}*{real prescription} & 
		\begin{CJK}{UTF8}{gbsn}
			甘草 法半夏 茯苓 前胡 桔梗 薏苡仁 浙贝母 细辛 天麻 鳖甲 款冬花 莪术 炙麻黄 蜈蚣 白花蛇舌
		\end{CJK}
		\\
		~                  & ~                                                &
		(Radix Glycyrrhizae, Rhizoma Pinellinae Praeparata, Poria Cocos, Radix Peucedani, Platycodonis Radix, Coicis Semen, Fritillaria Thunbergii Miq, Asarum Sieboldi Mig, Gastrodiae Rhizoma, Trionycis Carapax, Flos Farfarae, Curcuma Zedoary, Fried Herba Ephedrae, Scolopendra, Herba Hedyotidis)
		\\
		
		\cline{2-3}
		~                  & \multirow{2}*{conventional $CNN$}                 &
		\begin{CJK}{UTF8}{gbsn}
			\textcolor{red}{\textbf{甘草}} 柴胡 党参 酸枣仁 生地黄 红花 延胡索 浙贝母 山药 \textcolor{red}{\textbf{天麻 鳖甲 蜈蚣 白花蛇舌}} 天山雪莲 半枝莲
		\end{CJK}
		\\
		~                 & ~                                                 &
		(\textcolor{red}{\textbf{Radix Glycyrrhizae}}, Radix Bupleuri, Radix Codonopsis, Semen Zizyphi Spinosae, dried Rehamnnia Root, Carthamus Tinctorious, Corydalis Rhizoma, Thunberg Fritillary Bulb, Dioscoreae Rhizoma, \textcolor{red}{\textbf{Gastrodiae Rhizoma, Trionycis Carapax, Scolopendra, Herba Hedyotidis,}} Saussureae Involucratae Herba, Scutellariae Barbatae Herba)
		\\
		
		\cline{2-3}
		~                  & \multirow{2}*{conventional $CNN_{aug}$}           &
		\begin{CJK}{UTF8}{gbsn}
			\textcolor{red}{\textbf{甘草 茯苓}} 党参 白术 山药 \textcolor{red}{\textbf{天麻 鳖甲 蜈蚣 白花蛇舌}} 天山雪莲 半枝莲
		\end{CJK}
		\\
		~                 & ~                                                 &
		(\textcolor{red}{\textbf{Radix Glycyrrhizae, Poria Cocos,}} Radix Codonopsis, Macrocephalae Rhizoma, Dioscoreae Rhizoma, \textcolor{red}{\textbf{Gastrodiae Rhizoma, Trionycis Carapax, Scolopendra, Herba Hedyotidis,}} Saussureae Involucratae Herba, Scutellariae Barbatae Herba)
		\\
		
		\cline{2-3}
		~                  & \multirow{2}*{$CNN$ based on three-grained face}  &
		\begin{CJK}{UTF8}{gbsn}
			\textcolor{red}{\textbf{甘草 茯苓 薏苡仁}} 党参 \textcolor{red}{\textbf{天麻 鳖甲 蜈蚣 白花蛇舌}} 天山雪莲
		\end{CJK}
		\\
		~                  & ~                                                 &
		(\textcolor{red}{\textbf{Radix Glycyrrhizae, Poria Cocos, Coicis Semen,}} Radix Codonopsis, \textcolor{red}{\textbf{Gastrodiae Rhizoma, Trionycis Carapax, Scolopendra, Herba Hedyotidis,}} Saussureae Involucratae Herba)
		\\
		
		\cline{2-3}
		~                  & \multirow{2}*{$CNN_{aug}$ based on three-grained face}  &
		\begin{CJK}{UTF8}{gbsn}
			\textcolor{red}{\textbf{甘草 茯苓 薏苡仁}} 党参 浙贝母 \textcolor{red}{\textbf{天麻 鳖甲 蜈蚣 白花蛇舌}} 天山雪莲
		\end{CJK}
		\\
		~                  & ~                                                 &
		(\textcolor{red}{\textbf{Radix Glycyrrhizae, Poria Cocos, Coicis Semen,}} Radix Codonopsis, Fritillaria Thunbergii Miq, \textcolor{red}{\textbf{Gastrodiae Rhizoma, Trionycis Carapax, Scolopendra, Herba Hedyotidis,}} Saussureae Involucratae Herba)
		\\
		
		\midrule[2pt]
		\multirow{10}{*}[-80pt]{\raisebox{\height}{\includegraphics[width=50pt]{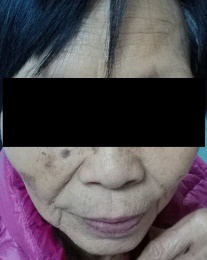}}} & \multirow{2}*{real prescription} & 
		\begin{CJK}{UTF8}{gbsn}
			法半夏 茯苓 前胡 桔梗 防风 白芷 款冬花 紫菀 白前 百部 炙甘草 辛夷 紫苏梗 广藿香 蜜麻黄
		\end{CJK}
		\\
		~                  & ~                                                &
		(Rhizoma Pinellinae Praeparata, Poria cocos, Radix Peucedani, Platycodonis Radix, Radix Saposhnikoviae, Radix Angelicae Dahuricae, Flos Farfarae, Aster Tataricus Linn,  Rhizoma Cynanchi Stauntonii, Radix Stemonae, Radix Glycyrrhizae Preparata, Flos Magnoliae, Perilla Stem, Herba Pogostemonis, Honey-fried Herba Ephedrae)
		\\
		
		\cline{2-3}
		~                  & \multirow{2}*{conventional $CNN$}                 &
		\begin{CJK}{UTF8}{gbsn}
			白芍 陈皮 \textcolor{red}{\textbf{防风}} 党参 枸杞子 \textcolor{red}{\textbf{白芷 炙甘草}} 首乌藤
		\end{CJK}
		\\
		~                 & ~                                                 &
		(Radix Paeoniae Alba, Tangerine Peel, \textcolor{red}{\textbf{Radix Saposhnikoviae,}} Radix Codonopsis, Fructus Lycii, \textcolor{red}{\textbf{Radix Angelicae Dahuricae, Radix Glycyrrhizae Preparata,}} Caulis Polygoni Multiflori)
		\\
		
		\cline{2-3}
		~                  & \multirow{2}*{conventional $CNN_{aug}$}           &
		\begin{CJK}{UTF8}{gbsn}
			麻黄 白芍 川芎 \textcolor{red}{\textbf{防风}} 荆芥穗 \textcolor{red}{\textbf{白芷}} 豆蔻 \textcolor{red}{\textbf{炙甘草 辛夷 广藿香}}
		\end{CJK}
		\\
		~                 & ~                                                 &
		(Herba Ephedrae, Radix Paeoniae Alba, Rhizoma Chuanxiong,\textcolor{red}{\textbf{ Radix Saposhnikoviae,}} Herba Schizonepetae, \textcolor{red}{\textbf{Radix Angelicae Dahuricae,}} Fructus Amomi Rotundus, \textcolor{red}{\textbf{Radix Glycyrrhizae Preparata, Flos Magnoliae, Herba Pogostemonis}})
		\\
		
		\cline{2-3}
		~                  & \multirow{2}*{$CNN$ based on three-grained face}  &
		\begin{CJK}{UTF8}{gbsn}
			陈皮 法半夏 \textcolor{red}{\textbf{茯苓 前胡}} 太子参 \textcolor{red}{\textbf{款冬花 紫菀}} 北沙参 \textcolor{red}{\textbf{炙甘草}}
		\end{CJK}
		\\
		~                  & ~                                                 &
		(Tangerine Peel, Rhizoma Pinellinae Praeparata, \textcolor{red}{\textbf{Poria Cocos, Radix Peucedani,}} Radix Pseudostellariae, \textcolor{red}{\textbf{Flos Farfarae, Aster Tataricus Linn,}} Radix Glehniae, \textcolor{red}{\textbf{Radix Glycyrrhizae Preparata}})
		\\
		
		\cline{2-3}
		~                  & \multirow{2}*{$CNN_{aug}$ based on three-grained face}  &
		\begin{CJK}{UTF8}{gbsn}
			\textcolor{red}{\textbf{法半夏 茯苓 前胡 桔梗 防风 款冬花 紫菀 白前 百部}} 北沙参 \textcolor{red}{\textbf{炙甘草 广藿香}} 炒紫苏子 \textcolor{red}{\textbf{蜜麻黄}}
		\end{CJK}
		\\
		~                  & ~                                                 &
		(\textcolor{red}{\textbf{Rhizoma Pinellinae Praeparata, Poria cocos, Radix Peucedani, Radix Platycodi, Radix Saposhnikoviae, Flos Farfarae, Aster Tataricus Linn, Rhizoma Cynanchi Stauntonii, Radix Stemonae,}} Radix Glehniae, \textcolor{red}{\textbf{Radix Glycyrrhizae Preparata, Herba Pogostemonis,}} Fried Perilla Fruit, \textcolor{red}{\textbf{Honey-fried Herba Ephedrae}})
		\\
		
		\bottomrule[2pt]
		\multicolumn{3}{p{500pt}}{The red bold type of Chinese herbal medicines indicate that it has appeared in the real prescription.}\\
		
	\end{tabular}
	\label{tab7}
\end{table*}

\subsubsection{Effect of different image size}
The input size of conventional $CNN$ is 224x224. The ``multi-scale'' of $CNN$ based on three-grained face means to the three-scale input 56x56, 112x112, 224x224, but the actual input size is still 224x224, which is the size of face image. We just input a 224x224 face image and the face image is segmented to 112x112 local region block images and 56x56 organ block images during preprocessing. So we say that the input size of $CNNs$ used above is 224x224. However, the size of the patient's face image in reality is uncertain. 

In order to verify that the CNN models can adapt to various images of different sizes, we retrain the networks of different input sizes with $D_{orgin}$ and get the experiment results, as shown in Table 6. The evaluation results(precision, recall, f1-score) are calculated by 5-fold cross validation methods. For multi-scale $CNN$ based on three-grained(we omit the ``multi-scale'' just for neat alignment in table 6), the image size means the size of face, and the size of local region block images is half of the size of face, the size of organ block images is half of the size of local region block images. The ``average'' in Table 6 means the average results of different image sizes(32, 56, 84, 112, 168, 224).

It can be seen from Table 6 that the models obtain similar results for different sizes of input images, indicating the robustness of the models. From the average, multi-scale $CNN$ based on three-grained face performs better than conventional $CNN$. In addition, for smaller image sizes(32, 56, 84), multi-scale $CNN$ based on three-grained face is slightly worse than conventional $CNN$, but when the input image is relatively large(112, 168, 224), multi-scale $CNN$ based on three-grained face is still excellent in the three evaluation indicators. We conjecture that multi-scale $CNN$ based on three-grained face needs to capture three-grained face features and fine-grained features are difficult to mine when the image is small. But on average, multi-scale $CNN$ based on three-grained face is still superior, and in reality it is unlikely to take too small patient's images. From the results of larger image sizes(112, 168, 224), the performances of multi-scale $CNN$ based on three-grained face are all higher than conventional $CNN$.

\section{Discussion}
Our results show that convolutional neural networks are capable of mining the prescription information from patient's face images to generate prescription, and the multi-scale convolutional neural networks based on three-grained indeed can generate prescriptions that are closer to real prescriptions, as shown in the actual prediction results in Table 7 and the evaluation results in Table 5 and Table 6. By building such a prescription generation system, the doctors can obtain recommended prescription, and then modify it, finally apply it to the actual treatment. 

Generation of TCM prescriptions from face image using deep learning can provide us with a possible result. Although the predicted result is not an inevitable conclusion, it provides us with a choice, a kind of opinion for reference, which greatly reduces the blindness of work. In fact, in reality, different TCM doctors do not always give the same prescriptions to patients, and there may be multiple prescriptions for the same patient. It is possible that system-generated prescriptions can inspire doctors to develop new useful prescriptions. 

\section{Conclusion}
In this paper, we propose to use convolutional neural network to generate TCM prescriptions according to the patient's face image. In order to more fully and effectively extract and utilize the features of the patient's face, we propose a multi-scale convolutional neural network based on three-grained face and compare it with the conventional convolutional neural network. In addition, we use data augmentation to increase the size and diversity of the data to improve the effect.

To the best of our knowledge, few people do the work to generate TCM prescriptions. Chinese herbal medicine is a medical asset accumulated by the Chinese ancient people’s long-term practice. It is extremely rich and precious. It is of great significance to fully mine and learn information from the prescribing data of patients using deep learning technique. 

In fact, when treating patients, doctors of TCM need to integrate multiple features (face, tongue, pulse, voice, symptoms) and their own experience to give solutions, which can overcome the limitations of using face images alone. Due to the limited data, in our preliminary research work we only consider to using patient’s face image to generate TCM prescriptions. In the future work, we plan to collect more quantities, more types of patient data.

\begin{acknowledgements}
This study was supported by the China National Science Foundation (60973083, 61273363), Science and Technology Planning Project of Guangdong Province (2014A010103009, 2015A020217002), and Guangzhou Science and Technology Planning Project (201504291154480, 2016040-
20179, 201803010088).
\end{acknowledgements}

\bibliographystyle{spbasic_unsort}      
\bibliography{reference}   

\begin{thebibliography}{45}
\providecommand{\natexlab}[1]{#1}
\providecommand{\url}[1]{{#1}}
\providecommand{\urlprefix}{URL }
\expandafter\ifx\csname urlstyle\endcsname\relax
  \providecommand{\doi}[1]{DOI~\discretionary{}{}{}#1}\else
  \providecommand{\doi}{DOI~\discretionary{}{}{}\begingroup
  \urlstyle{rm}\Url}\fi
\providecommand{\eprint}[2][]{\url{#2}}

\bibitem[{Cheung(2011)}]{Cheung2011Tcm}
Cheung F (2011) {TCM: Made in China}. Nature 480:S82

\bibitem[{Qiu(2007)}]{Qiu2007TraditionalMedicine}
Qiu J (2007) {Traditional medicine: A culture in the balance}. Nature 448:126

\bibitem[{{H. Peng}(1996)}]{Peng1996DictOfTcm}
{H Peng} (1996) {Dictionary of Traditional Chinese Medicine Prescriptions}.
  People Health Press: Beijing,China

\bibitem[{Yao et~al.(2018)Yao, Zhang, Wei, Zhang, and
  Jin}]{Yao2018TcmTopicModel}
Yao L, Zhang Y, Wei B, Zhang W, Jin Z (2018) {A Topic Modeling Approach for
  Traditional Chinese Medicine Prescriptions}. IEEE Transactions on Knowledge
  and Data Engineering 30(6):1007--1021

\bibitem[{Yiqin(2012)}]{Wang2012TcmInspection}
Yiqin W (2012) {Objective Application of TCM Inspection of Face And Tongue}.
  Chinese Archives of Traditional Chinese Medicine 30(2):349--352

\bibitem[{Jones(2018)}]{Jones2018FaceHealth}
Jones AL (2018) {The influence of shape and colour cue classes on facial health
  perception}. Evolution and Human Behavior 39(1):19--29

\bibitem[{Chen et~al.(2017)Chen, Shi, Zhang, Wu, and
  Guizani}]{Chen2017DeepLearningForMedical}
Chen M, Shi X, Zhang Y, Wu D, Guizani M (2017) {Deep Features Learning for
  Medical Image Analysis with Convolutional Autoencoder Neural Network}. IEEE
  Transactions on Big Data p~1

\bibitem[{Litjens et~al.(2017)Litjens, Kooi, Bejnordi, Setio, Ciompi,
  Ghafoorian, van~der Laak, van Ginneken, and
  S{\'{a}}nchez}]{Litjens2017Survey}
Litjens G, Kooi T, Bejnordi BE, Setio AAA, Ciompi F, Ghafoorian M, van~der Laak
  JAWM, van Ginneken B, S{\'{a}}nchez CI (2017) {A survey on deep learning in
  medical image analysis}. Medical Image Analysis 42:60--88

\bibitem[{Bayramoglu et~al.(2016)Bayramoglu, Kannala, and
  Heikkil{\"{a}}}]{Bayramoglu2016BreastCancerClassify}
Bayramoglu N, Kannala J, Heikkil{\"{a}} J (2016) {Deep learning for
  magnification independent breast cancer histopathology image classification}.
  In: 2016 23rd International Conference on Pattern Recognition (ICPR), pp
  2440--2445

\bibitem[{Chougrad et~al.(2018)Chougrad, Zouaki, and
  Alheyane}]{Chougrad2018BreastCancerScreening}
Chougrad H, Zouaki H, Alheyane O (2018) {Deep Convolutional Neural Networks for
  breast cancer screening}. Computer Methods and Programs in Biomedicine
  157:19--30

\bibitem[{Anthimopoulos et~al.(2016)Anthimopoulos, Christodoulidis, Ebner,
  Christe, and Mougiakakou}]{Anthimopoulos2016LungClassify}
Anthimopoulos M, Christodoulidis S, Ebner L, Christe A, Mougiakakou S (2016)
  {Lung Pattern Classification for Interstitial Lung Diseases Using a Deep
  Convolutional Neural Network}. IEEE Transactions on Medical Imaging
  35(5):1207--1216

\bibitem[{Hon and Khan(2017)}]{Hon2017AlzheimerClassify}
Hon M, Khan NM (2017) {Towards Alzheimer's disease classification through
  transfer learning}. In: 2017 IEEE International Conference on Bioinformatics
  and Biomedicine (BIBM), pp 1166--1169

\bibitem[{Stanitsas et~al.(2017)Stanitsas, Cherian, Truskinovsky, Morellas, and
  Papanikolopoulos}]{Stanitsas2017CancerRecognize}
Stanitsas P, Cherian A, Truskinovsky A, Morellas V, Papanikolopoulos N (2017)
  {Active convolutional neural networks for cancerous tissue recognition}. In:
  2017 IEEE International Conference on Image Processing (ICIP), pp 1367--1371

\bibitem[{Wang et~al.(2017)Wang, Ding, Bidgoli, Zhou, Iribarren, Molloi, and
  Baldi}]{Wang2017DetectCardiovascular}
Wang J, Ding H, Bidgoli FA, Zhou B, Iribarren C, Molloi S, Baldi P (2017)
  {Detecting Cardiovascular Disease from Mammograms With Deep Learning}. IEEE
  Transactions on Medical Imaging 36(5):1172--1181

\bibitem[{Yu et~al.(2017)Yu, Chen, Dou, Qin, and
  Heng}]{Yu2017MelanomaRecognize}
Yu L, Chen H, Dou Q, Qin J, Heng PA (2017) {Automated Melanoma Recognition in
  Dermoscopy Images via Very Deep Residual Networks}. IEEE Transactions on
  Medical Imaging 36(4):994--1004

\bibitem[{Yuan et~al.(2017)Yuan, Chao, and Lo}]{Yuan2017SkinSegment}
Yuan Y, Chao M, Lo YC (2017) {Automatic Skin Lesion Segmentation Using Deep
  Fully Convolutional Networks With Jaccard Distance}. IEEE Transactions on
  Medical Imaging 36(9):1876--1886

\bibitem[{Kassim et~al.(2017)Kassim, Prasath, Glinskii, Glinsky, Huxley, and
  Palaniappan}]{Kassim2017MicrovasculatureSegment}
Kassim YM, Prasath VBS, Glinskii OV, Glinsky VV, Huxley VH, Palaniappan K
  (2017) {Microvasculature segmentation of arterioles using deep CNN}. In: 2017
  IEEE International Conference on Image Processing (ICIP), pp 580--584

\bibitem[{Zhao et~al.(2018)Zhao, Wu, Song, Li, Zhang, and
  Fan}]{Zhao2018BrainTumourSegmantation}
Zhao X, Wu Y, Song G, Li Z, Zhang Y, Fan Y (2018) {A deep learning model
  integrating FCNNs and CRFs for brain tumor segmentation}. Medical Image
  Analysis 43:98--111

\bibitem[{Chaabouni et~al.(2017)Chaabouni, Benois-pineau, Tison, Ben~Amar, and
  Zemmari}]{Chaabouni2017}
Chaabouni S, Benois-pineau J, Tison F, Ben~Amar C, Zemmari A (2017) Prediction
  of visual attention with deep cnn on artificially degraded videos for studies
  of attention of patients with dementia. Multimedia Tools and Applications
  76(21):22527--22546

\bibitem[{Xu et~al.(2017)Xu, Liu, Cheng, Song, and
  Zhang}]{Xu2017CirrhosisDiagnosis}
Xu Z, Liu X, Cheng XE, Song JL, Zhang JQ (2017) {Diagnosis of cirrhosis stage
  via deep neural network}. In: 2017 IEEE International Conference on
  Bioinformatics and Biomedicine (BIBM), pp 745--749

\bibitem[{Zhao et~al.(2017)Zhao, Dong, Chen, Iraji, Li, Makkie, Kou, and
  Liu}]{Zhao2017BrainNetworkAtlases}
Zhao Y, Dong Q, Chen H, Iraji A, Li Y, Makkie M, Kou Z, Liu T (2017)
  {Constructing fine-granularity functional brain network atlases via deep
  convolutional autoencoder}. Medical Image Analysis 42:200--211

\bibitem[{Liu et~al.(2012)Liu, Zhou, Wang, Hu, He, Zhang, Chen, and
  Guo}]{Liu2012TcmDataProcess}
Liu B, Zhou X, Wang Y, Hu J, He L, Zhang R, Chen S, Guo Y (2012) {Data
  processing and analysis in real‐world traditional Chinese medicine clinical
  data: challenges and approaches}. Statistics in Medicine 31(7):653--660

\bibitem[{Zheng et~al.(2014)Zheng, Jiang, Lu, and Lu}]{Zheng2014Prescription}
Zheng G, Jiang M, Lu C, Lu A (2014) {Prescription Analysis and Mining},
  Springer International Publishing, Cham, pp 97--109

\bibitem[{Xie et~al.(2017)Xie, Pei, Zhu, and Li}]{Xie2017Prescription}
Xie D, Pei W, Zhu W, Li X (2017) {Traditional Chinese medicine prescription
  mining based on abstract text}. In: 2017 IEEE 19th International Conference
  on e-Health Networking, Applications and Services (Healthcom), pp 1--5

\bibitem[{Zhang et~al.(2012)Zhang, Zhang, and Chen}]{Zhang2012Prescription}
Zhang NL, Zhang R, Chen T (2012) {Discovery of Regularities in the Use of Herbs
  in Traditional Chinese Medicine Prescriptions}. In: Cao L, Huang JZ, Bailey
  J, Koh YS, Luo J (eds) New Frontiers in Applied Data Mining, Springer Berlin
  Heidelberg, Berlin, Heidelberg, pp 353--360

\bibitem[{Yao et~al.(2015)Yao, Zhang, Wei, Wang, Zhang, Ren, and
  Bian}]{Yao2015TcmTopicModel}
Yao L, Zhang Y, Wei B, Wang W, Zhang Y, Ren X, Bian Y (2015) {Discovering
  treatment pattern in Traditional Chinese Medicine clinical cases by
  exploiting supervised topic model and domain knowledge}. Journal of
  Biomedical Informatics 58:260--267

\bibitem[{Dehan et~al.(2014)Dehan, Jia, Yimin, and
  Hamid}]{Dehan2014TcmMedicineClassify}
Dehan L, Jia W, Yimin C, Hamid G (2014) {Classification of Chinese Herbal
  medicines based on SVM}. In: 2014 International Conference on Information
  Science, Electronics and Electrical Engineering, vol~1, pp 453--456

\bibitem[{Weng et~al.(2017)Weng, Hu, and Lan}]{Weng2017TcmRecognize}
Weng JC, Hu MC, Lan KC (2017) {Recognition of Easily-confused TCM Herbs Using
  Deep Learning}. In: Proceedings of the 8th ACM on Multimedia Systems
  Conference, ACM, New York, NY, USA, MMSys'17, pp 233--234

\bibitem[{Yu et~al.(2017)Yu, Li, Yu, Tian, Shun, Xu, Zhu, and
  Gao}]{Yu2017TcmKnowledgeGraph}
Yu T, Li J, Yu Q, Tian Y, Shun X, Xu L, Zhu L, Gao H (2017) {Knowledge graph
  for TCM health preservation: Design, construction, and applications}.
  Artificial Intelligence in Medicine 77:48--52

\bibitem[{Weng et~al.(2017)Weng, Liu, Yan, Fan, Ou, Chen, and
  Hao}]{Weng2017TcmKnowledgeGraph}
Weng H, Liu Z, Yan S, Fan M, Ou A, Chen D, Hao T (2017) {A Framework for
  Automated Knowledge Graph Construction Towards Traditional Chinese Medicine}.
  In: Siuly S, Huang Z, Aickelin U, Zhou R, Wang H, Zhang Y, Klimenko S (eds)
  Health Information Science, Springer International Publishing, Cham, pp
  170--181

\bibitem[{Zhang and Zhou(2014)}]{Zhang2014MultiLabel}
Zhang ML, Zhou ZH (2014) {A Review on Multi-Label Learning Algorithms}. IEEE
  Transactions on Knowledge and Data Engineering 26(8):1819--1837

\bibitem[{Krizhevsky et~al.(2012)Krizhevsky, Sutskever, and
  Hinton}]{Krizhevsky2012AlexNet}
Krizhevsky A, Sutskever I, Hinton GE (2012) {ImageNet Classification with Deep
  Convolutional Neural Networks}. Advances In Neural Information Processing
  Systems pp 1097--1105, \eprint{1102.0183}

\bibitem[{Simonyan and Zisserman(2015)}]{Simonyan2015VGG}
Simonyan K, Zisserman A (2015) {Very Deep Convolutional Networks for
  Large-Scale Image Recognition}. International Conference on Learning
  Representations (ICRL) \eprint{1409.1556}

\bibitem[{Szegedy et~al.(2015)Szegedy, Liu, Jia, Sermanet, Reed, Anguelov,
  Erhan, Vanhoucke, and Rabinovich}]{Szegedy2015GoogleNet}
Szegedy C, Liu W, Jia Y, Sermanet P, Reed S, Anguelov D, Erhan D, Vanhoucke V,
  Rabinovich A (2015) {Going deeper with convolutions}. In: 2015 IEEE
  Conference on Computer Vision and Pattern Recognition (CVPR), pp 1--9

\bibitem[{He et~al.(2016)He, Zhang, Ren, and Sun}]{He2016ResNet}
He K, Zhang X, Ren S, Sun J (2016) {Deep Residual Learning for Image
  Recognition}. In: 2016 IEEE Conference on Computer Vision and Pattern
  Recognition (CVPR), pp 770--778

\bibitem[{Xie et~al.(2017)Xie, Girshick, Doll{\'{a}}r, Tu, and
  He}]{Xie2017ResNeXt}
Xie S, Girshick R, Doll{\'{a}}r P, Tu Z, He K (2017) {Aggregated Residual
  Transformations for Deep Neural Networks}. In: 2017 IEEE Conference on
  Computer Vision and Pattern Recognition (CVPR), pp 5987--5995

\bibitem[{Zagoruyko and Komodakis(2016)}]{Zagoruyko2016wrn}
Zagoruyko S, Komodakis N (2016) {Wide Residual Networks}. In: {Richard C
  Wilson} ERH, Smith WAP (eds) Proceedings of the British Machine Vision
  Conference (BMVC), BMVA Press, pp 87.1--87.12

\bibitem[{Huang et~al.(2017)Huang, Liu, v.~d. Maaten, and
  Weinberger}]{Huang2017DenseNet}
Huang G, Liu Z, v~d Maaten L, Weinberger KQ (2017) {Densely Connected
  Convolutional Networks}. In: 2017 IEEE Conference on Computer Vision and
  Pattern Recognition (CVPR), pp 2261--2269

\bibitem[{Hu et~al.(2018)Hu, Shen, and Sun}]{Hu2018seNet}
Hu J, Shen L, Sun G (2018) {Squeeze-and-Excitation Networks}. In: The IEEE
  Conference on Computer Vision and Pattern Recognition (CVPR)

\bibitem[{Glorot et~al.(2011)Glorot, Bordes, and Bengio}]{Glorot2011Relu}
Glorot X, Bordes A, Bengio Y (2011) {Deep sparse rectifier neural networks}.
  In: AISTATS '11: Proceedings of the 14th International Conference on
  Artificial Intelligence and Statistics, pp 315--323, \eprint{1502.03167}

\bibitem[{Bottou(2012)}]{Bottou2012SGD}
Bottou L (2012) {Stochastic Gradient Descent Tricks}. In: Neural Networks:
  Tricks of the Trade, Springer, Berlin, Heidelberg, pp 421--436,
  \eprint{9780201398298}

\bibitem[{Taigman et~al.(2014)Taigman, Yang, Ranzato, and
  Wolf}]{Taigman2014DeepFace}
Taigman Y, Yang M, Ranzato M, Wolf L (2014) {DeepFace: Closing the Gap to
  Human-Level Performance in Face Verification}. In: 2014 IEEE Conference on
  Computer Vision and Pattern Recognition, pp 1701--1708

\bibitem[{Jain and Learned-Miller(2010)}]{Vidit2010fddb}
Jain V, Learned-Miller E (2010) Fddb: A benchmark for face detection in
  unconstrained settings. Tech. Rep. UM-CS-2010-009, University of
  Massachusetts, Amherst

\bibitem[{Srivastava et~al.(2014)Srivastava, Hinton, Krizhevsky, Sutskever, and
  Salakhutdinov}]{Srivastava2014Dropout}
Srivastava N, Hinton G, Krizhevsky A, Sutskever I, Salakhutdinov R (2014)
  {Dropout: A Simple Way to Prevent Neural Networks from Overfitting}. Journal
  of Machine Learning Research 15(1):1929--1958, \eprint{1102.4807}

\bibitem[{Breiman(2001)}]{Breiman2001RandomForest}
Breiman L (2001) {Random Forests}. Machine Learning 45(1):5--32

\end{thebibliography}

\end{document}